\documentclass[final]{cvpr}

\usepackage{times}
\usepackage{epsfig}
\usepackage{graphicx}
\usepackage{amsmath}
\usepackage{amssymb}
\usepackage{textcomp}
\usepackage[font=small]{caption}
\usepackage[caption=false,font=scriptsize]{subfig}
\usepackage{hhline}
\usepackage{enumitem}
\usepackage{boldline}
\usepackage{comment}

\setlist{itemsep=2pt,parsep=2pt}

\newlength{\tempheight}
\newlength{\tempwidth}

\newcommand{\rowname}[1]
{\rotatebox{90}{\makebox[\tempheight][c]{\small \textbf{#1}}}}

\newcommand{\columnname}[1]
{\makebox[\tempwidth][c]{\textbf{#1}}}

\usepackage[pagebackref=true,breaklinks=true,colorlinks,bookmarks=false]{hyperref}

\def\cvprPaperID{6376} 
\def\confYear{CVPR 2021}

\begin{document}

\title{No Shadow Left Behind: Removing Objects and their Shadows using Approximate Lighting and Geometry}

\author{
    Edward Zhang\textsuperscript{1}
    \and Ricardo Martin-Brualla\textsuperscript{2}
    \and Janne Kontkanen\textsuperscript{2}
    \and Brian Curless\textsuperscript{1,2} \smallskip
    \and
    \textsuperscript{1}University of Washington\qquad
    \textsuperscript{2}Google
}

\twocolumn[{%
\renewcommand\twocolumn[1][]{#1}%
\maketitle
    \begin{center}

    \centering
\includegraphics[width=0.99\textwidth]{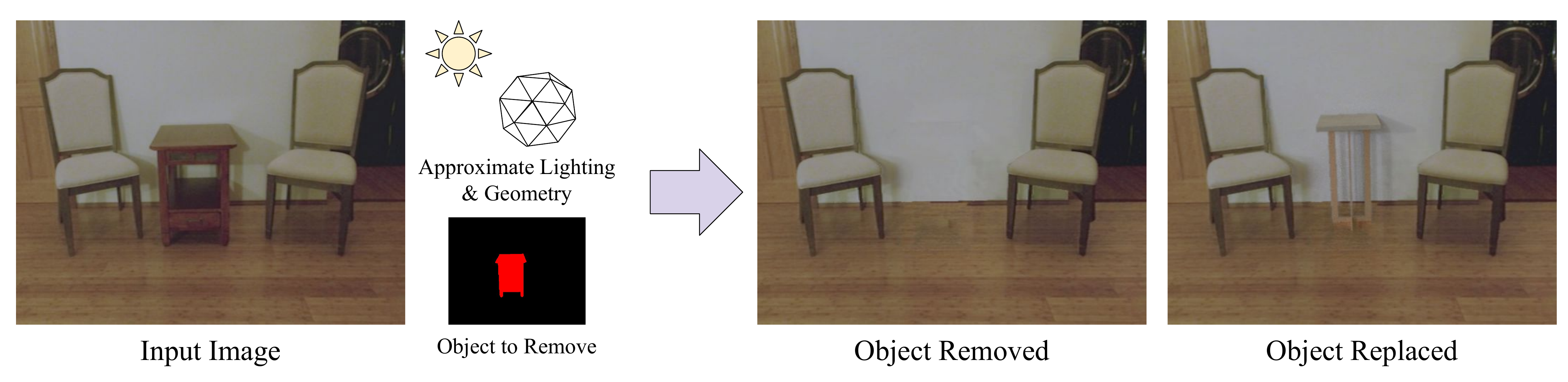}
        \captionof{figure}{
        We present a method to remove an object and its shadows from an image, to enable applications like home refurnishing. Our method takes as input an image, approximate scene lighting and geometry, and an object mask, and generates a new version of the image that depicts the scene as if the object had not been present. This not only includes inpainting the occluded pixels, but removing any shadows cast by the object.
}
\label{fig:teaser}
    \end{center}
}]

\begin{abstract}
    Removing objects from images is a challenging technical problem that is
    important for many applications, including mixed reality.
    For believable results, the shadows that the object casts should also be
    removed. Current inpainting-based methods only remove the object itself,
    leaving shadows behind, or at best require specifying shadow regions to inpaint.
    We introduce a deep learning pipeline for removing a
    shadow along with its caster. We leverage rough scene models in order to remove
    a wide variety of shadows (hard or soft, dark or subtle, large or thin) from
    surfaces with a wide variety of textures. We train our pipeline on
    synthetically rendered data, and show qualitative and quantitative results on
    both synthetic and real scenes.
\end{abstract}

\section{Introduction}
Mixed reality aims to seamlessly combine the virtual and the real.
As one example, imagine an interior design app that lets you try out new furniture.
Most previous work in augmented reality focuses on inserting virtual objects
 -- for instance, putting a virtual sofa into your
living room.
The scope of these applications can be greatly expanded by also enabling manipulation of 
real-world objects -- imagine removing the futon that you intend to
replace with the sofa, and moving a coffee table over to make more room for it.

Previous work on object removal has focused solely on the inpainting problem --
that is, replacing the pixels previously occupied by the removed object.
However, for realistic results, we need to remove the sofa {\it and} the shadows
it casts on the wall and the floor, as well as the reflection on the hardwood
floor. For the purposes of this paper, we focus only on the shadow removal problem.

Existing inpainting-based approaches for object removal either ignore the shadows
of the object, or mark them to be inpainted as well. However, very large shadows may leave little image content to
copy pixels from. Furthermore, this approach requires segmenting out the object's shadow in addition to the object itself -- a difficult task,
as varying lighting conditions can cause multiple shadows, very soft
shadows, or overlapping shadows, and a surface texture may have dark regions that could be mistaken for shadows.

Inspired by Debevec's \cite{debevec1998rendering} work in virtual insertion of
objects in scenes, we use a scene proxy to help determine
the visual effects of a scene manipulation. Debevec performs the
scene edit on the proxy model, and renders the proxy pre- and post-edit. The pixelwise difference between the two renderings, which for
object insertion contains shadows and reflections of the virtual objects, is
then applied to the input image to produce the final output. This
method is known as {\it differential rendering}.
However, it is not practical to solve the shadow removal problem by applying
the pixelwise difference directly, since the shadows in the proxy model are only
a rough estimate of the real shadows.
To account for this, we propose a neural network based system for more general
differential rendering for object removal.

An obvious question is, how do we obtain an editable scene proxy? One could use a depth camera, monocular depth estimation~\cite{monodepth, monodepth2}, or a global model obtained as a side effect of localization~\cite{bundlefusion} for the geometry. For lighting, the possibilities include a mirror sphere, panorama, or learning based methods~\cite{legendre2019deeplight, song2019neural, srinivasan2020lighthouse}. In this paper we use depth maps captured by an affordable depth sensor and a $360^{\circ}$  panorama, but the method is not fundamentally limited to proxies obtained by these devices, nor do the proxy models need to be very accurate. Our proxy mesh is generated from a single depth map, thus modeling only front facing surfaces, and our lighting is captured as an uncalibrated HDR environment map with only very rough alignment. We show that even this constrained and incomplete proxy provides enough information to generate plausible removal results across a wide range of conditions.

In this paper we present a method for removing an object and its shadows from
an input image, given a rough model of the scene and the mask of the object. Our
system is more accurate and produces fewer visual artifacts than a general
image-to-image translation system or an inpainting method, even when the
inpainting method is given the shadow regions it should replace.

\begin{figure*}[ht!]
    \centering
\includegraphics[width=0.98\textwidth]{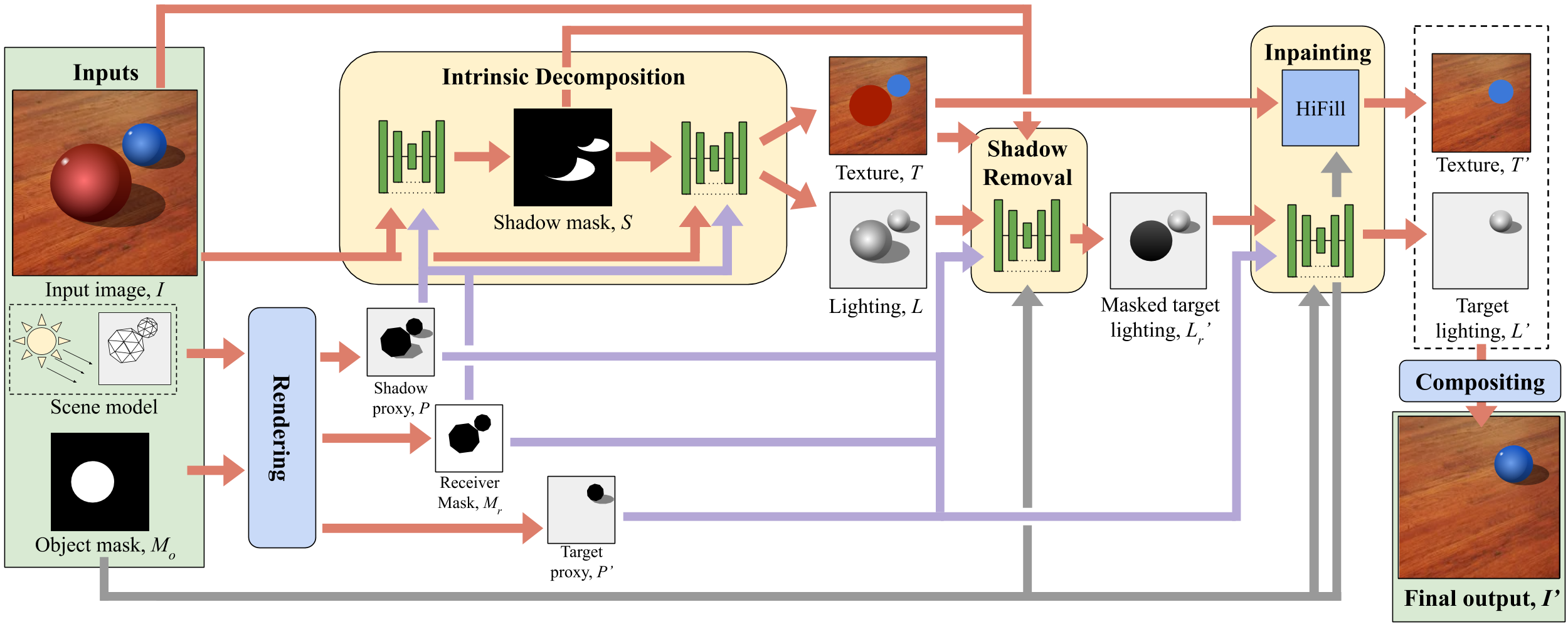}
    \caption{Overview of our approach. Our system takes an input image, a mask of the
    object to remove, and renderings of an approximate scene model with
    and without the object. We first perform an intrinsic decomposition of the 
    input image into texture and lighting. We then remove the object's shadows 
    from the lighting image. We inpaint the object mask region separately in
    both texture and lighting before recompositing to get our final
    result.}
\label{fig:architecture}
\end{figure*}

\section{Related Work}
\subsection{Scene Editing}
Editing scenes in a visually realistic manner has long been an area of interest
in the graphics community.
Most of this work has focused on virtual object insertion. Classical
methods construct an approximate model of the scene to help perform these edits,
ranging from Debevec's early work \cite{debevec1998rendering}, which assumes
lighting and geometry were directly captured, to more recent work by Karsch et
al. \cite{karsch2014automatic}, which infers geometry, albedo, and lighting
from a single image. Beyond simply inserting objects, Kholgade \etal \cite{kholgade20143d} are able to move an object around, although they
assume that a high-quality 3D model of the object is available.

Research on object removal has traditionally focused on the
inpainting problem, ranging from classical techniques such as
PatchMatch \cite{patchmatch} to recent learning-based techniques such as
DeepFill \cite{deepfill} and HiFill \cite{hifill}. These methods do not consider
lighting interactions between the removed object and the rest of the scene;
thus when removing the object by inpainting, the user-specified mask must
be extended to include the object's shadow. Recent
work by Wang \etal \cite{wang2020instance} employs deep networks to associate shadows with their
casters; however, their instance segmentation approach produces hard
boundaries and does not work for soft shadows. Zhang \etal \cite{zhang2016emptying} remove objects from indoor scenes
by constructing a full scene model and rendering it without the
objects, eliminating the need for inpainting and shadow identification;
however, their approach requires an involved capture process and is limited by the expressivity of their parametric scene model.

The issue of the limited range of scene models is inherent to all of the methods
that rely on such models for scene editing. Most works (e.g. Kholgade \cite{kholgade20143d}) use Debevec's differential rendering
method to account for differences between the model and the real scene. Recent approaches for neural rerendering
use image-to-image translation to map from the domain of the
approximate model to a realistic result \cite{lookingood, meshry2019neural}.
Philip \etal \cite{philip2019multiview} introduced a method for relighting outdoor photographs that also
leverages proxy scene models. Their system is
designed to handle global changes in illumination, like changing the position of
the sun; furthermore they rely heavily on shadow masks that cannot handle complex
environment lighting. Instead, we focus on local changes informed by the
differences in the appearance of the proxy scene, based on an intrinsic
decomposition that can handle multiple soft shadows.


\subsection{Shadow Removal}
Removing shadows from images is another problem that has a long history. Note
that the goal of these works is to remove all shadows from an image, while
our goal is to remove the shadow of a single object; however, in our work we
assume the presence of a rough scene proxy.

Classical intrinsic image decomposition methods are designed with various priors, typically specializing in low-frequency lighting and thus handling
soft shadows well \cite{sirfs, scenesirfs, iiw, grosse2009ground}. Another set
of methods specialize in hard shadows and classify gradients as shading or texture
\cite{bell2001learning, tappen2005recovering}; however, these methods break down
when shadow receivers have complex texture.
Finlayson \etal \cite{finlayson2005removal} place assumptions on light source chromaticity, allowing
for removal of both soft and hard shadows at the expense of generality.

Recent methods use deep networks to perform shadow detection
and removal, starting with work by Qu \etal \cite{srd}. Advances such as
adversarial losses \cite{istd, ding2019argan},
a two-stage detection-then-removal scheme \cite{hu2019direction},
or lighting inference \cite{le2019shadow} have resulted in great
improvements on shadow removal on the common ISTD \cite{istd} and SRD \cite{srd}
datasets. However, these datasets only contain hard shadows produced from
outdoor lighting. Our system is trained to handle much more diversity in lighting conditions.

\section{Method}

Our pipeline consists of a series of convolutional neural networks (we
use a single U-Net \cite{unet} for each component),
with an inpainting stage to produce our final results. A visual overview
can be seen in Figure~\ref{fig:architecture}. Our system takes as inputs the
original image, a rendering of the approximate scene model before object
removal (referred to as the {\it shadow proxy}), a rendering of
the scene model after object removal ({\it target proxy}), and
binary masks denoting the object to remove and the receiving surface from which to
remove shadows. An overview of our pipeline follows:

\begin{itemize}
\item The {\bf intrinsic decomposition subsystem} separates the input image into texture and lighting, guided by
        the shadow proxy. Following existing works \cite{hu2019direction} on shadow removal, we use a two-stage
scheme with an initial shadow segmentation network.
\item The {\bf shadow removal} network removes the shadow of the removed object from the decomposed
    lighting, aided by the shadow proxy and target proxy images.
\item The {\bf inpainting subsystem} separately inpaints the lighting and texture behind the removed object. Our learned lighting inpainting uses the target proxy to inform where the remaining
shadows in the scene should continue behind the removed object. For texture inpainting, we use an off-the-shelf inpainting method. Inpainting the decomposed texture, rather than the final composite,
prevents the inpainting method from hallucinating its own shadows.
\item Lastly, we recompose the lighting and texture images back together to
    produce our final result.
\end{itemize}

We define $I, I'$ as the input and output images, respectively. $P, P'$ are the shadow proxy and target
proxy. The intrinsic decomposition is denoted by $I = LT, I' = L'T'$ with $L,L'$
being the lighting images and $T,T'$ being the reflectance (texture) images.
$M_o$ is a binary mask which is 1 for pixels lying on the object to be removed and 0 elsewhere. $M_r$ is a binary
receiver mask which is 1 for pixels lying on the local scene receiving the shadows, and 0 for pixels elsewhere. This restricts the network to operate on a surface with a single texture, as otherwise the intrinsic decomposition frequently mislabels shadows as changes in reflectance if the surface reflectance can vary arbitrarily.

RGB images are processed in the log domain, turning the intrinsic decomposition
$I = LT$ into a sum $\log(I) = \log(L) + \log(T)$ that is more naturally
represented by CNNs. Using synthetic training data enables full
supervision of each subnetwork's intermediate outputs.

{\bf Shadow Segmentation:} This subnetwork produces a 1-channel soft
segmentation in the range $[0, 1]$ with 1 denoting full shadow.

\begin{equation}
{S} = f_\text{SS}(\log(I), \log(P), M_r)
\end{equation}

{\bf Intrinsic Decomposition:} This subnetwork decomposes the input image into two 3-channel
outputs: lighting ${L}$ and texture ${T}$.
\begin{equation}
{L}, {T} = \exp(f_\text{ID}(\log(I), \log(P), {S}, M_r))
\end{equation}

{\bf Shadow Removal:} This subnetwork removes the shadow of the
object from the predicted lighting image, producing one 3-channel output, the
masked target lighting.
\begin{equation}
\begin{aligned}
{L}_r' = \exp(f_\text{SR}(&\log(I),\log({T}),\log({L}),\\  & \log(P), \log(P'), {S}, M_r, M_o))
\end{aligned}
\end{equation}

{\bf Lighting Inpainting:} This subnetwork fills in the predicted lighting
${L}_r'$ behind the removed object, continuing shadows cast by other objects through the mask if
necessary. The target lighting ${L}'$ is then the composite of the inpainted
lighting and the masked target lighting.
\begin{equation}
{L}_o' = \exp(f_\text{LI}(\log({L}_r'), \log(P'), M_o))
\end{equation}
\begin{equation}
{L}' = (1-M_o){L}_r' + M_o{L}_o'
\end{equation}

{\bf Texture Inpainting:}
We inpaint the texture image using an inpainting
operator $g(T,M_o)$, synthesizing the pixels of $T$ in the hole region specified
by the mask image $M_o$. For our experiments we used
 HiFill \cite{hifill} for $g(T,M_o)$, trained on the Places2 dataset
 \cite{places2}.
\begin{equation}
{T}' = g({T}, M_o)
\end{equation}

{\bf Final Composite:}
The previous stages predicted the appearance of the receiver within the target receiver mask
$M'_r = M_r + M_o$. We composite the remaining pixels from the
original image, consisting of unaffected surfaces beyond the local scene and other
objects within the local scene.

\begin{equation}
{I}' = M'_r{T}'{L}' + (1-M'_r)I
\end{equation}

\subsection{Training}
Each subnetwork is independently trained with the Adam optimizer for 60 epochs
on 60000 training scenes,
with ground truth intermediates substituted for the outputs of earlier
subnetworks with a learning rate of $10^{-4}$ decaying by $0.5$ every 10 epochs.
We then train the whole system end-to-end for 60 epochs. Our networks were
implemented in Tensorflow and trained on four Tesla V100 GPUs with a batch size of 16.

Our loss functions are described below.
For all networks except the lighting inpainting network, the inputs to the
losses are masked with $M_r$ to only apply to pixels lying on the
receiver. For brevity we assume that the norm flattens
across input channels and image dimensions. We denote a ground truth supervision
image with a hat, so that the ground truth intrinsic decomposition is $\hat{L},
\hat{T}$, ground truth output image is $\hat{I}'$, and so on.

{\bf Shadow Segmentation:}
It is difficult to define a ground truth for what constitutes a shadow in a
scene lit by an HDRI map, since any object will occlude some part of the
distant illumination. To supervise this
stage we therefore define a shadow as any pixel where the ground
truth lighting is less than the median pixel intensity on any of the three
color channels using a soft threshold:

\begin{equation}
\hat{S} = \max\left(\sigma \left(\frac{\text{median}(\hat{L})-\hat{L}}{\alpha} \right)\right)
\end{equation}

The shadow segmentation subnetwork is supervised by a class-balanced binary
cross entropy term as well as a loss on the gradients of the shadow
segmentation:

\begin{equation}
E_\text{SS} = \lambda_SE_{S} + \lambda_{\nabla S}E_{\nabla S}
\end{equation}
\begin{equation}
E_{S} = \frac{-\hat{S}\log({S})}{||\hat{S}||_1} -
\frac{(1-\hat{S})\log(1-{S})}{||1-\hat{S}||_1}
\end{equation}
\begin{equation}
E_{\nabla S} = ||\nabla S - \nabla \hat{S}||_2
\end{equation}

{\bf Intrinsic Decomposition:}
The intrinsic decomposition loss function is the most involved of our losses, 
including a data term, two terms for the decomposition and a data prior.
\begin{equation}
E_\text{ID} = \lambda_{LT}E_{LT} + \lambda_\text{excl}E_\text{excl} +
\lambda_IE_{I} + \lambda_{\nabla L}E_{\nabla L}
\end{equation}
For the data term, a multiscale loss on the predicted lighting and texture images
was vital to ensure the model would work well on high-contrast textures.
\begin{equation}
E_{LT} = P(L, \hat{L}) + P(T, \hat{T})
\end{equation}
where $P(X,\hat{X})$ is an L2 loss on a Gaussian pyramid decomposition of the images
$X,\hat{X}$.

To ensure a clean decomposition, we impose the exclusion losses of Zhang \etal \cite{zhang2018single} on the predicted lighting and
texture images, which in
essence constructs 0-to-1-valued edge maps at multiple scales, and penalizes edges lying at the same location in the two
decomposed images.
\begin{equation}
E_\text{excl} = \sum_{i=0}^{i=3} 4^i||\Psi({T} \downarrow
i,{L} \downarrow i)||
\end{equation}

where $X \downarrow n$ denotes image $X$ downsampled bilinearly by a factor of
$2^n$, and $\Psi$ is as defined by Zhang et al.

We also have an L1 loss on the two decomposed images recomposing into the input image.
\begin{equation}
E_{I} = ||I - LT||_1
\end{equation}
Finally, we impose a sparse gradient prior on $L$ to discourage textural details from leaking into the lighting.
\begin{equation}
E_{\nabla L} = ||\nabla L||_1
\end{equation}

{\bf Shadow Removal and Lighting Inpainting:}
As with the intrinsic decomposition, we apply a multiscale loss on the
predicted lighting after shadow removal and inpainting.

\begin{equation}
E_{L'} = \lambda_{L'}P(L', \hat{L}')
\end{equation}

Note that because $L'$ is a composite of the results of the shadow removal
and lighting inpainting networks, the shadow removal network is only penalized
for pixels lying on the receiver in the original input image while the
lighting inpainting network is only penalized for pixels within
the mask of the removed object.

We also add a recomposition loss on the final output.

\begin{equation}
E_{I'} = \lambda_{I'}||\hat{I}' - L'T'||_1
\end{equation}
\section{Training Data}
Capturing a large set of paired images with and without a particular object would
require a prohibitive amount of labor. Therefore, we use a synthetic
dataset to train our system, which also allows us to generate the ground truth for the intermediate stages such as the intrinsic decomposition.

To generate training data, we set up 60000 input scenes with randomly generated
geometry, textures, lighting, and camera parameters. These scenes are rendered
using PBRT \cite{pbrt} to produce images of resolution $512 \times 512$.
The dataset exhibits a wide variety of shadow casters (e.g large objects, thin structures, and objects with unusual silhouettes) and lighting conditions (hard or soft shadows, very dark or very
subtle shadows, multiple shadows).

\subsection{Scene Generation}
{\bf Geometry: }
Our generated scenes consist of a ground plane supporting six to seven objects
randomly selected from the \texttildelow50000 3D models in the ShapeNet
\cite{shapenet} dataset, which include a variety of object classes ranging
from furniture and tableware to cars and airplanes. These objects are
arranged in a ring around a central object, and are scaled such that the
bounding boxes are nonintersecting. Each object is translated such that it lies
entirely on top of the plane, and has a random rotation around its up axis. The
ground plane is large enough to support all the shadow casters, plus an
additional margin for shadows to potentially fall upon.

{\bf Materials:}
The supporting plane is given a matte material, and is assigned a random
texture (e.g. carpet, wood, stone, tile). Existing texture datasets are too
small (e.g. the Brodatz textures \cite{brodatz}) or have textures which are nonuniform (e.g. the Describable
Textures Dataset \cite{dtd}).
We use a manually curated texture dataset of about 8000 images from Google Image
Search results.
The ShapeNet objects come with prespecified materials.

{\bf Lighting:}
We illuminate each scene by one of the \texttildelow400
HDRI maps at HDRI Haven\footnote{https://hdrihaven.com/}, randomly rotated around the up axis. To supplement the
lighting, we add a point light with random intensity
(setting the maximum to the peak intensity of the HDRI map) randomly placed between a minimum
and maximum distance from the center of the plane, in the upper hemisphere.

{\bf Camera:}
We define a range of camera positions lying on an upper hemisphere of
fixed radius in terms of spherical coordinates facing the center of the scene. We allow the azimuthal angle to
vary freely, but set a minimum and maximum elevation angle (as people rarely
observe scenes from directly overhead or from very low angles). After selecting
an initial camera pose we then perturb the camera's position while keeping the
same orientation.

\subsection{Image Generation}
\label{sec:trainingdata:imagegeneration}

Using PBRT,
we render three RGB images of each scene: $\hat{T}$, the ground plane alone with diffuse texture; $\hat{L}$, the complete scene with the plane
material replaced by a diffuse white material; and $\hat{L}'$, the scene with the
central object removed with the same alteration to the plane material.
These images comprise the ground truth
intrinsic decomposition's texture and lighting, and the lighting
post-object-removal. Note that these images do not form a true intrinsic decomposition of the entire scene, only of the receiving plane.

\begin{table*}[t]
    \centering
    \footnotesize
    \begin{tabular}{|c|c|c|c||c|c|c|}
        \hline
        & \multicolumn{3}{c||}{Synthetic} & \multicolumn{3}{c|}{Real}\\ \hline
                                   & RMSE   & Shadow RMSE  & Inpaint RMSE & RMSE         & Shadow RMSE  & Inpaint RMSE \\ \hline
        No-op                      & 0.0455 & 0.2785       & 0.3755       & 0.0405       & 0.1413       & 0.2702 \\ \hlineB{3}
        PatchMatch                 & 0.0460 & 0.2790       & 0.2565       & 0.0401       & 0.1378       & 0.1288 \\ \hline
        HiFill                     & 0.0479 & 0.2700       & 0.2555       & 0.0408       & 0.1305       & 0.1160 \\ \hlineB{3}
        PatchMatch + Shadows       & 0.0402 & 0.2143       & 0.2282       & {\bf 0.0351} & 0.0969       & 0.1025 \\ \hline
        HiFill + Shadows           & 0.0461 & 0.2193       & 0.2346       & 0.0365       & 0.0855       & {\bf 0.0993} \\ \hlineB{3}
        Pix2Pix (all)              & 0.3583 & 0.3243       & 0.4477       & 0.2502       & 0.2649       & 0.3129 \\ \hline
        Pix2Pix (receiver)         & 0.0820 & 0.2118       & 0.2323       & 0.1091       & 0.1526       & 0.1266     \\ \hlineB{3}
        Pix2Pix + Proxy (receiver) & 0.0802 & 0.1766       & 0.2244       & 0.0872       & 0.1187       & 0.1153 \\ \hline     
        +Intrinsic Decomposition         & 0.0362 & 0.0882 & 0.2238 & 0.0618 & 0.0843 & 0.1055 \\ \hline
        +Shadow Segmentation             & {\bf 0.0246} & {\bf 0.0713} & 0.2213 & {\bf 0.0340} & 0.0631 & 0.1040 \\ \hline
        +Lighting Inpainting (Ours)      & {\bf 0.0248} & {\bf 0.0712} & {\bf 0.2143} & {\bf 0.0340} & {\bf 0.0616} & {\bf 0.0983}\\ \hline
    \end{tabular}
    \caption{Comparison of error rates for various shadow removal methods}
    \label{table:end2end}
\end{table*}

Next, we render depth maps $D, D'$ of the unedited and target scenes, as well as a
depth map $D_r$ of solely the ground plane receiving shadows, all using the same camera pose as the RGB images. From these we compute a pixel mask of the
object to be removed $M_o = \mathbb{I}(D' < D)$ which is 1 where the object is and 0
everywhere else. We also compute a receiver mask $M_r = \mathbb{I}(D_r \ne \infty, D_r = D)$
which is 1 for pixels lying on the ground plane in the unedited scene and 0
everywhere else.

To allow for further augmentation, we do not raytrace the input
and output images $I,\hat{I}'$, instead computing them at train time from the decomposition: $I =
(M_r\hat{T} + (1-M_r)) \hat{L}$ and $\hat{I}' = (M'_r\hat{T} + (1-M'_r))\hat{L}'$.
This allows us to modify the hue, saturation, and brightness of texture and
lighting at train time.
Note that we forgo indirect bounce lighting in our synthetic data to enable this
augmentation, as indirect illumination depends on the surface reflectance, \ie texture.

To mimic real capture, we add noise to the depth map of the unedited scene and construct a triangle mesh
from the depth map as our approximate geometry, replacing the ground plane vertices with a best-fit
plane (which continues behind the removed object). To form the target proxy
geometry, we delete the depth pixels occupied by the removed object. This scene is lit with a perturbed
version of the input lighting: we jitter the point light's position, color, and intensity, apply a random nonlinear scale to the HDRI map, and randomly
rotate the HDRI map by a small amount. All materials are set to a diffuse white; note that we do not model the surface reflectance
(texture) of the plane as it would imply already knowing the intrinsic decomposition.
Rendering these elements produces $P, P'$, respectively the images of
the unedited and target scene proxy.

\subsection{Normalization}
Our intrinsic decomposition has a scale ambiguity, which we resolve by
normalizing the ground truth lighting $\hat{L}, \hat{L}'$ at train
time, expecting that the network will produce normalized lighting images.
Specifically, we apply a per-channel scale to both $\hat{L}, \hat{L}'$ such that the maximum
pixel value on the receiver across both images is $(1,1,1)$.

Similarly, we compute a scale factor to normalize the images of the
scene proxy $P, P'$ (which are just approximations of $\hat{L}, \hat{L}'$).
This occurs at both test and train time.

For both train and test we scale the images in the input domain (i.e.
$I$ and $\hat{I}'$) to have a channelwise mean pixel value of 0.5 on the ground plane.



\section{Results}
\begin{figure*}
\setlength{\tempwidth}{.18\linewidth}
\settoheight{\tempheight}{\includegraphics[width=\tempwidth]{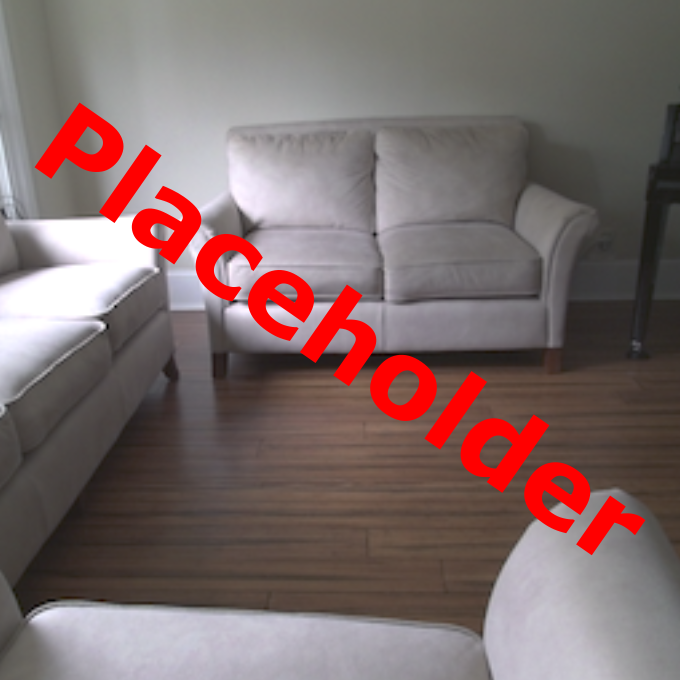}}%
\centering
\hspace{\baselineskip}
\columnname{Input Image}\hfil
\columnname{Ground Truth}\hfil
\columnname{Pix2Pix+Proxy}\hfil
\columnname{HiFill+Shadows}\hfil
\columnname{Ours}\hfil \\
\rowname{a. Synthetic 1}
\includegraphics[width=\tempwidth]{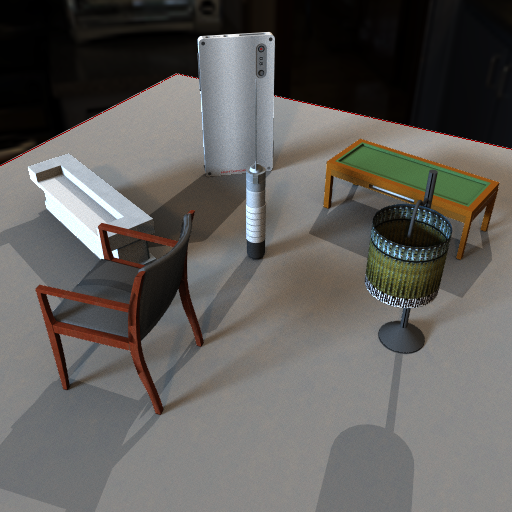}~
\includegraphics[width=\tempwidth]{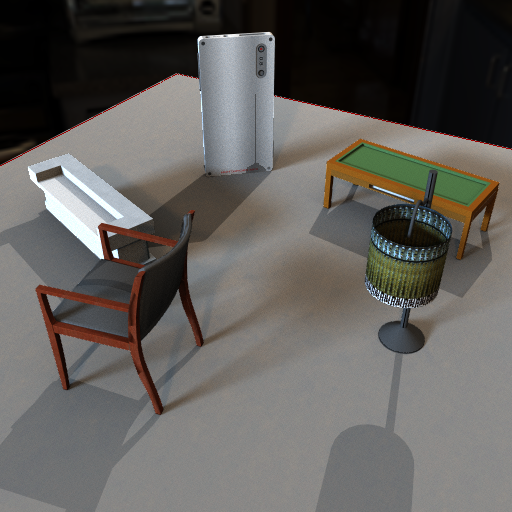}~
\includegraphics[width=\tempwidth]{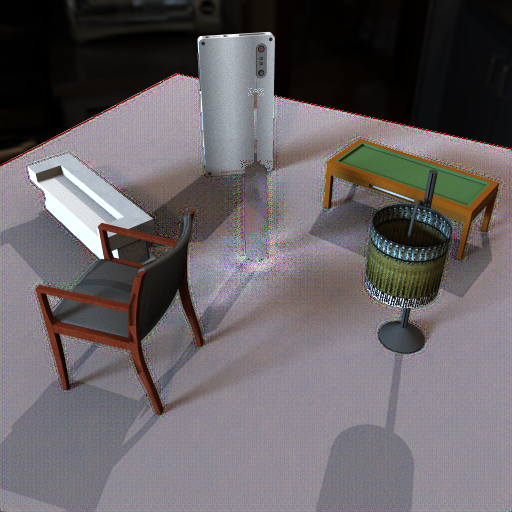}~
\includegraphics[width=\tempwidth]{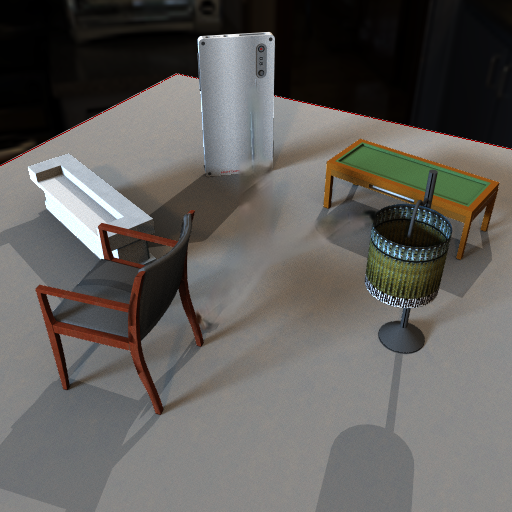}~
    \includegraphics[width=\tempwidth]{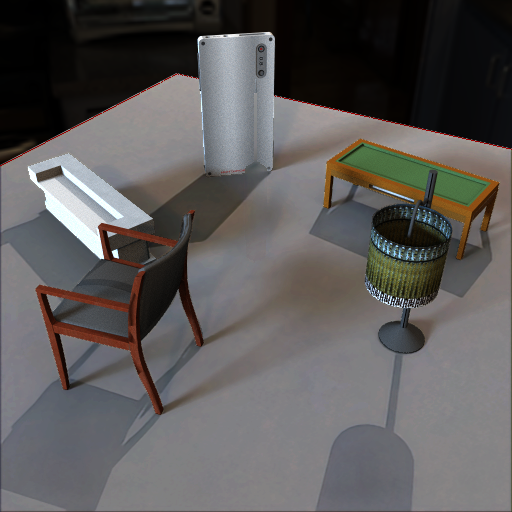}\\[0.5ex]
\rowname{b. Synthetic 2}
\includegraphics[width=\tempwidth]{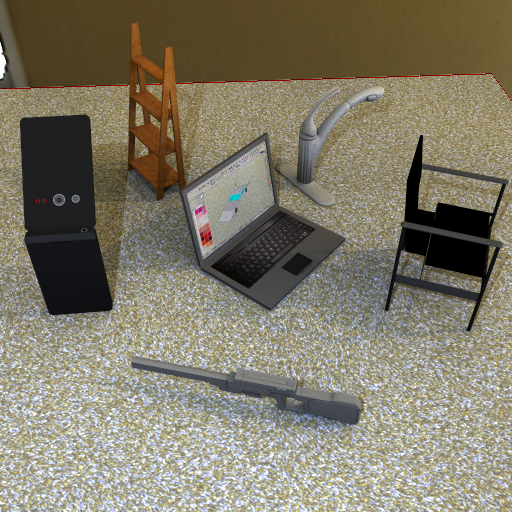}~
\includegraphics[width=\tempwidth]{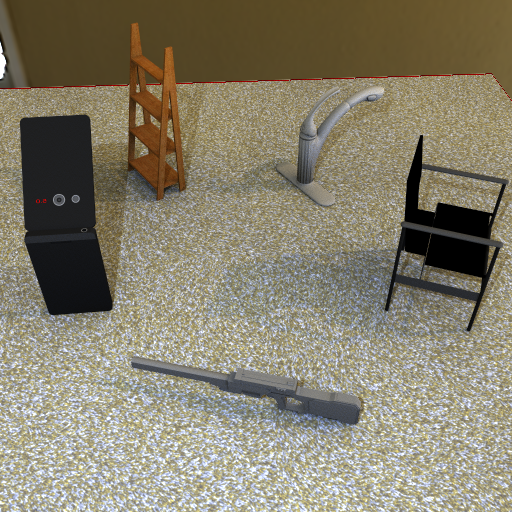}~
\includegraphics[width=\tempwidth]{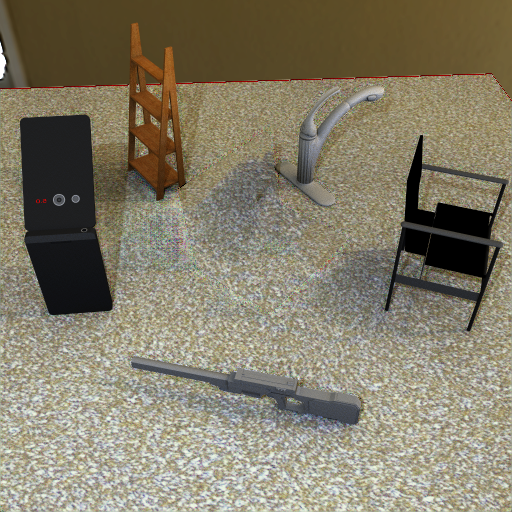}~
\includegraphics[width=\tempwidth]{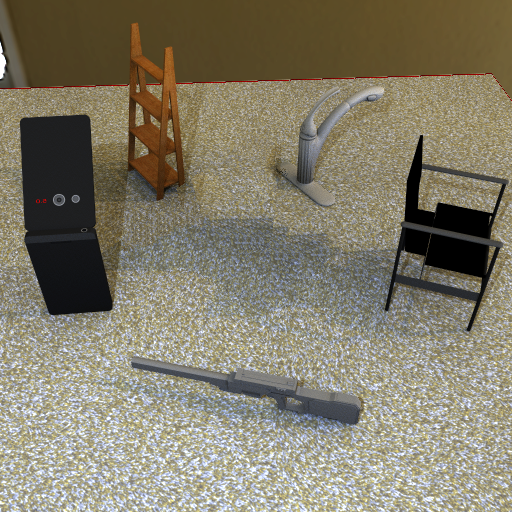}~
    \includegraphics[width=\tempwidth]{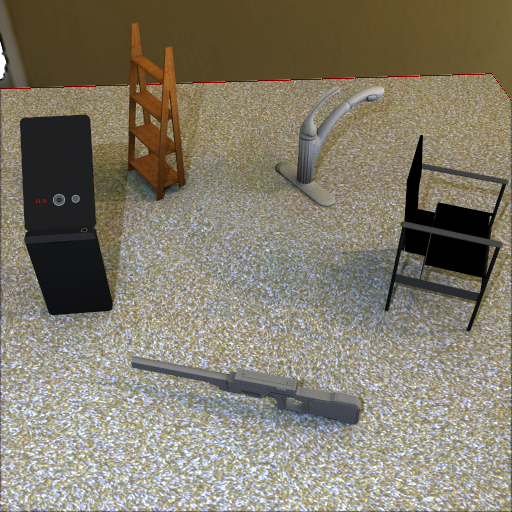}\\[0.5ex]
\settoheight{\tempheight}{\includegraphics[width=0.7\tempwidth]{figures/placeholder.png}}%
\rowname{c. Real 1}
\includegraphics[width=\tempwidth,height=\tempheight]{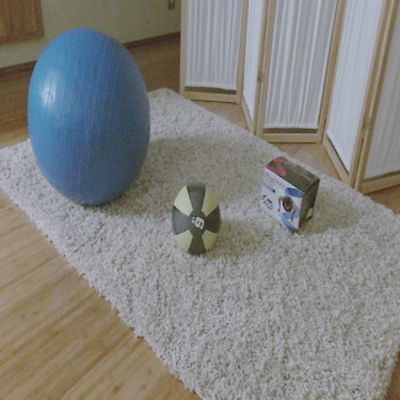}~
\includegraphics[width=\tempwidth,height=\tempheight]{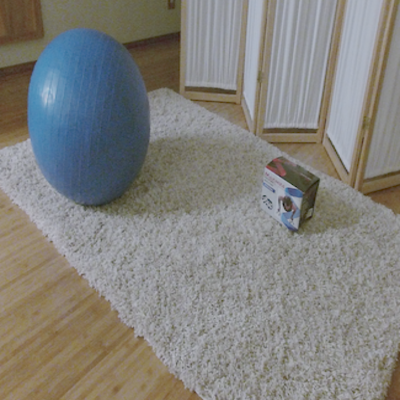}~
\includegraphics[width=\tempwidth,height=\tempheight]{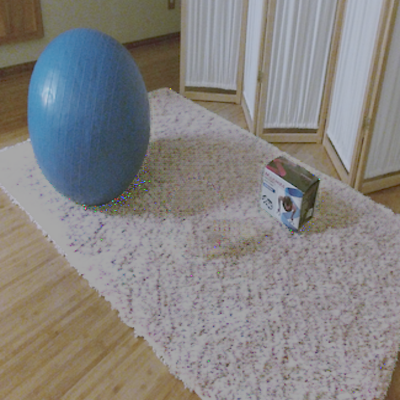}~
\includegraphics[width=\tempwidth,height=\tempheight]{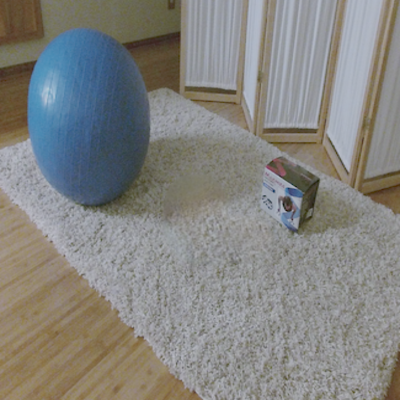}~
    \includegraphics[width=\tempwidth,height=\tempheight]{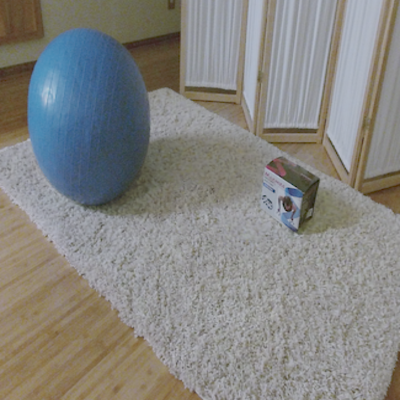}\\[0.5ex]
\rowname{d. Real 2}
\includegraphics[width=\tempwidth,height=\tempheight]{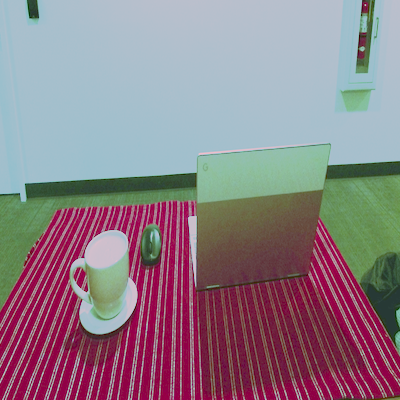}~
\includegraphics[width=\tempwidth,height=\tempheight]{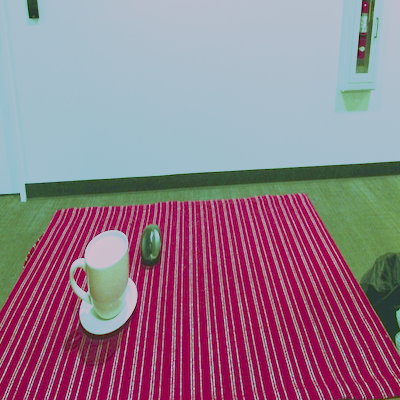}~
\includegraphics[width=\tempwidth,height=\tempheight]{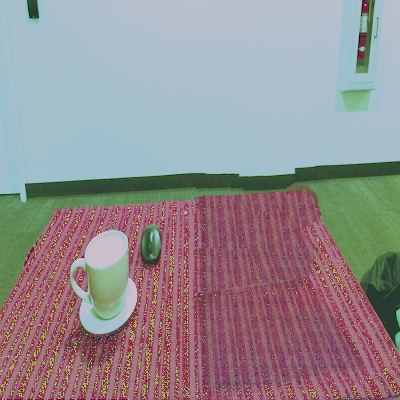}~
\includegraphics[width=\tempwidth,height=\tempheight]{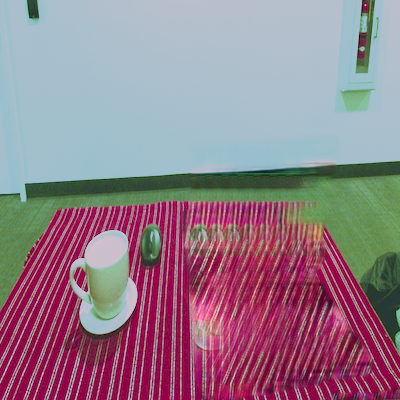}~
    \includegraphics[width=\tempwidth,height=\tempheight]{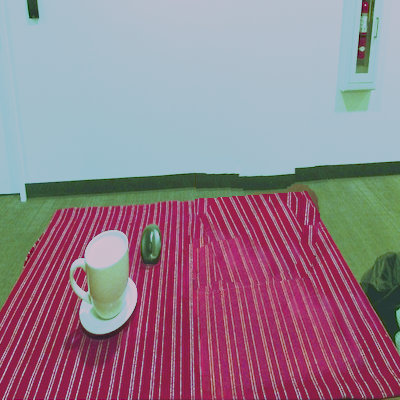}\\[0.5ex]
\rowname{e. Real 3}
\includegraphics[width=\tempwidth,height=\tempheight]{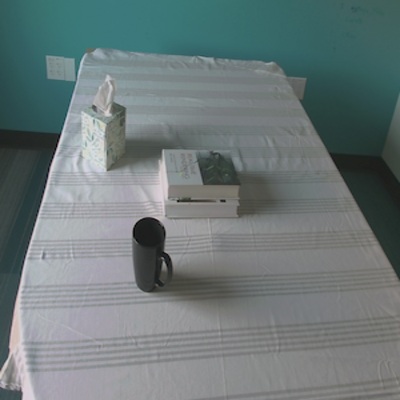}~
\includegraphics[width=\tempwidth,height=\tempheight]{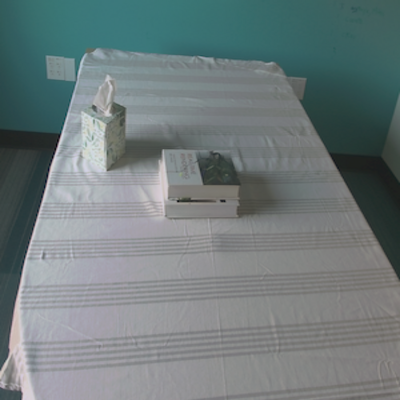}~
\includegraphics[width=\tempwidth,height=\tempheight]{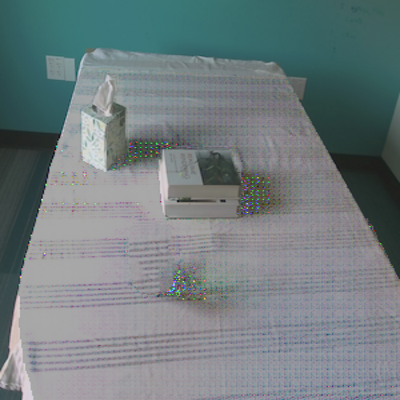}~
\includegraphics[width=\tempwidth,height=\tempheight]{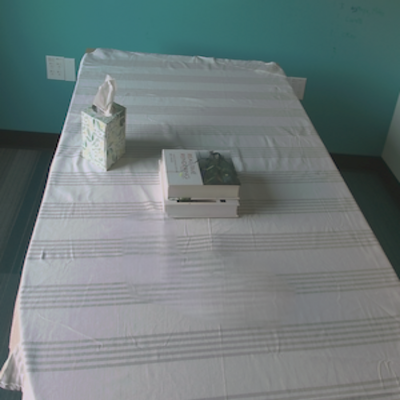}~
\includegraphics[width=\tempwidth,height=\tempheight]{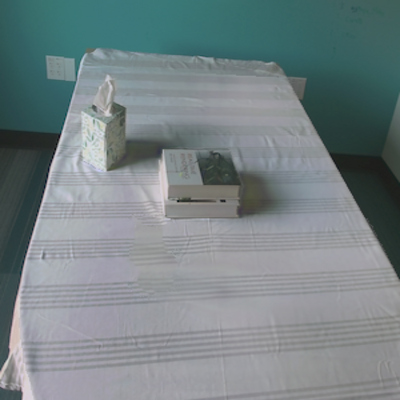}
\caption{We compare our system to two object removal baselines on both synthetic and real test
    images. The first baseline is an image-to-image translation network based on 
    Pix2Pix \cite{pix2pix} which is supplied with our renderings of the proxy 
    scene. The second baseline is HiFill \cite{hifill}, a state-of-the-art inpainting method,
    that inpaints both the removed object and an explicitly specified shadow mask.}
\label{fig:end2end}
\end{figure*}

We evaluate our work both qualitatively and quantitatively on 5000 synthetic test
scenes, generated the same way as our training data, and 14 real
scenes captured manually, which include ground truth object removal results.

For quantitative results, we report three separate RMSE metrics. The basic RMSE
measured across all pixels in the scene is a poor representation of the quality
of shadow removal results. A perceptually negligible color cast produced by a
deep network across the entire image has an outsized effect on the RMSE. To
better represent the performance of various systems, we also report the Shadow
RMSE, which is computed across pixels within the ground truth binary shadow mask
$\hat{S}$. Finally, we separately report the RMSE within the removed object's pixels.


\subsection{Test Data}
For the real scenes, the proxy geometry and the images were captured using the RGBD Kinect v2 device mounted on a tripod. We captured three frames for each scene: $I$ the complete scene, $\hat{I}'$ the target image with one or more objects removed, and a bare scene with all objects removed.
The approximate lighting was captured using a Ricoh Theta S 360 camera with 5 exposures for high dynamic range placed approximately in the center of the scene, roughly pointed at the Kinect.

We median-filtered the depth images to remove noise. We then
computed the best-fit plane for the input depth image
using RANSAC \cite{ransac}, and computed the receiver mask $M_r$ to indicate pixels which
approximately lie on the plane. For this work we manually specified the object masks
$M_o$; in real applications a more sophisticated automatic
segmentation method could be used.

To compute the proxy geometry, we formed a triangle mesh from the depth map as we did with the
synthetic data; however for cleaner shadows we replaced the vertices
corresponding to the ground plane with a fitted plane. This geometry was then
rendered with the captured HDR environment lighting to produce $P,P'$. As we did
not have ground truth intrinsic decompositions, we used the difference
between $I$ and $\hat{I}'$ instead of $\hat{L}$ and $\hat{L}'$ to produce the
shadow mask $\hat{S}$ used in the Shadow RMSE metric.

\subsection{End-to-End Comparisons}
Most existing works on object removal do not focus on
removing shadows cast by the object. We compare to two general approaches as baselines: pure inpainting
and generic image-to-image translation. We also include the numerical error of the
``no-op'' procedure, which does not transform the input at all.
Quantitative results are shown in Table~\ref{table:end2end} and
qualitative results in Figure~\ref{fig:end2end}, exhibiting varying lighting
conditions (multiple overlapping shadows, soft and hard shadows, high
and low contrast shadows) and background textures in both synthetic and
real test scenes.

We compute the inpainting baselines using two methods: the classical
nonparametric PatchMatch \cite{patchmatch}, and a recent learning-based
approach HiFill \cite{hifill}. For both baselines, we
include quantitative results both for a naive hole-fill, where the object's shadows are
not handled at all, as well as for an inpainting mask which includes the object's
shadows. In Figure~\ref{fig:end2end} we show the results of HiFill inpainting when
provided the shadow region. With this extra information, simple inpainting approaches
work well for simple textures (\ref{fig:end2end}a,\ref{fig:end2end}c), but often
hallucinate shadows within the inpainted region (\ref{fig:end2end}b). They also fail to take
advantage of texture detail inside the shadow region, and the resulting artifacts are compounded
when the region to inpaint is large (\ref{fig:end2end}d).

\begin{figure*}[ht!]

    \centering
\subfloat[Input]{\includegraphics[width=0.16\textwidth]{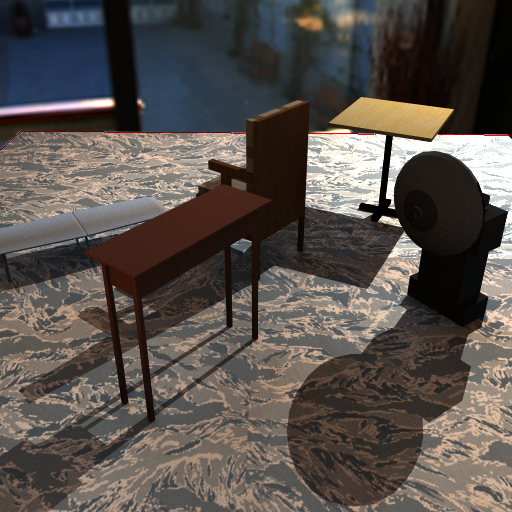} \label{fig:ablation:input}}~
\subfloat[Ground Truth]{\includegraphics[width=0.16\textwidth]{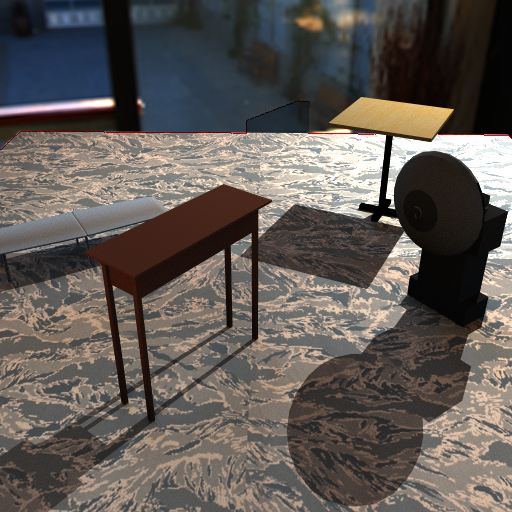} \label{fig:ablation:gt}}~
\subfloat[Pix2pix (e.g. U-net)]{\includegraphics[width=0.16\textwidth]{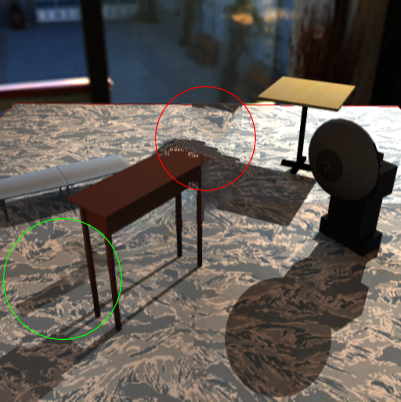} \label{fig:ablation:1}}~
\subfloat[+Intrinsic Decomposition]{\includegraphics[width=0.16\textwidth]{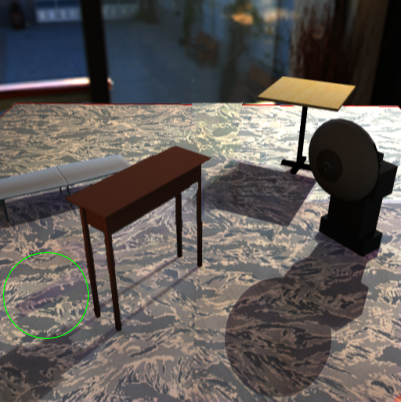} \label{fig:ablation:2}}~
\subfloat[+Shadow Segmentation]{\includegraphics[width=0.16\textwidth]{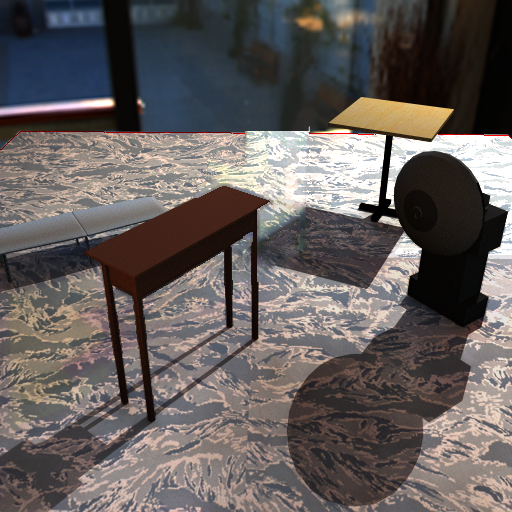} \label{fig:ablation:3}}~
\subfloat[+Lighting Inpainting]{\includegraphics[width=0.16\textwidth]{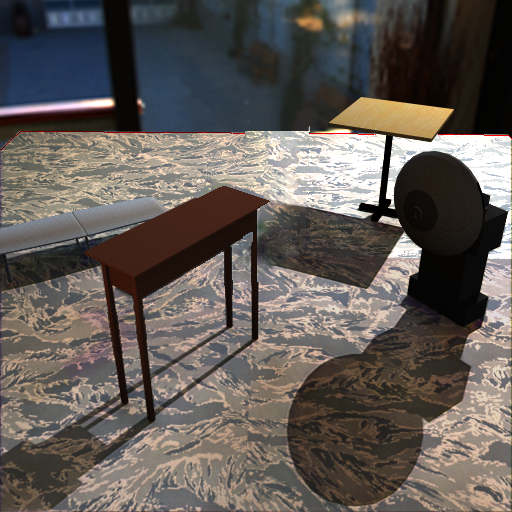} \label{fig:ablation:4}}
\caption{This synthetic example (a) shows the importance of each component of our system. The ground truth removal is shown in (b). A single U-net (c) hallucinates shadows in the inpainted region (red) and misidentifies shadows (green) with high-contrast textures. Adding in a texture decomposition subnetwork significantly improves the inpainting (d); however the edges of the shadows are still faintly visible (green). The shadow segmentation subnetwork eliminates these shadow ghosts (e). Some final artifacts visible within the silhouette of the removed object are fixed with our lighting inpainting network (f).}
\label{fig:ablation}
\end{figure*}

For our image-to-image translation baseline, we use the well-known Pix2Pix
method \cite{pix2pix}. We compare against three variants of this
baseline: one trained to predict the entire output image $I'$ from the input
$I$
and object mask $M_o$; one trained to predict only the appearance of the
receiving surface (using HiFill to inpaint the object region);
and one trained to predict only the appearance of the receiver but supplied
with our proxy scene renderings $P, P'$ in addition to the input image and
object mask. We show the results of this last version in
Figure~\ref{fig:end2end}. The method fails to accurately identify the extents
of shadows (\ref{fig:end2end}c,\ref{fig:end2end}e) and their intensities (\ref{fig:end2end}a)
and generalizes poorly to complex textures (\ref{fig:end2end}b,\ref{fig:end2end}d).

\subsection{Validating our Architecture}
We show the importance of each step of our pipeline, starting with a single network to perform our generalized differential rendering task and adding in each component one by one. The results are shown in Figure~\ref{fig:ablation} and the bottom rows of Table~\ref{table:end2end}.
We start with a single Pix2pix U-Net generator, that given all our inputs, directly predicts
the output image excluding the pixels under the object mask, which are inpainted
using HiFill (Figure~\ref{fig:ablation:1}). The most obvious artifacts are within the inpainted region, where the inpainting method frequently fills in shadow pixels; this method also frequently misidentifies shadows, especially in high-contrast textures.
We then introduce a separate intrinsic decomposition network,
and allow the shadow removal network to work only on the resulting
lighting image (Figure~\ref{fig:ablation:2}); however sometimes this system does not completely remove hard or high-contrast shadows. Introducing a shadow segmentation network (Figure~\ref{fig:ablation:3}) makes decompositions of hard shadows much cleaner. Finally, we introduce a dedicated lighting inpainting network (Figure~\ref{fig:ablation:4}), as the shadow removal network alone has trouble continuing shadows behind the object and sometimes leaves visible artifacts in the hole region.

\begin{figure}[h!]
    \centering
\includegraphics[width=0.48\columnwidth,height=0.33\columnwidth]{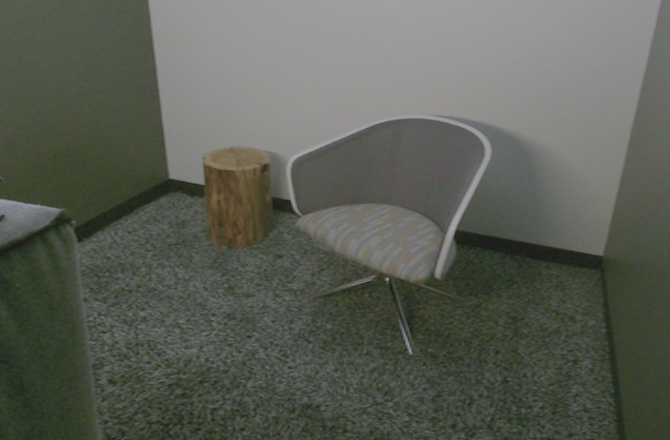}~
    \includegraphics[width=0.48\columnwidth,height=0.33\columnwidth]{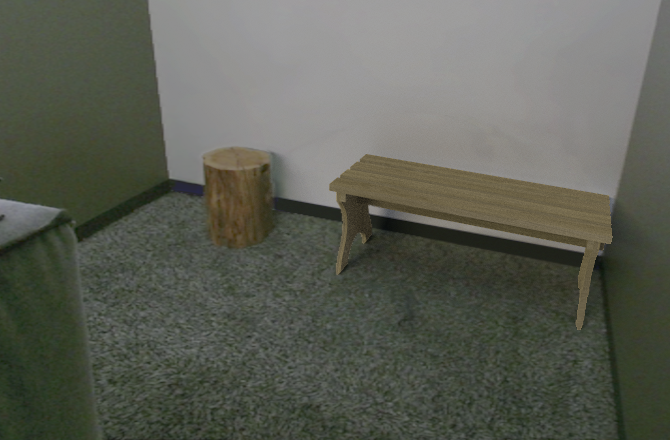}\\[0.5ex]
\includegraphics[width=0.48\columnwidth,height=0.3\columnwidth]{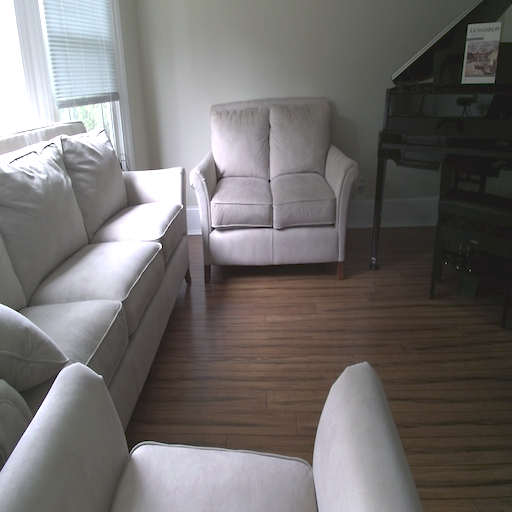}~
\includegraphics[width=0.48\columnwidth,height=0.3\columnwidth]{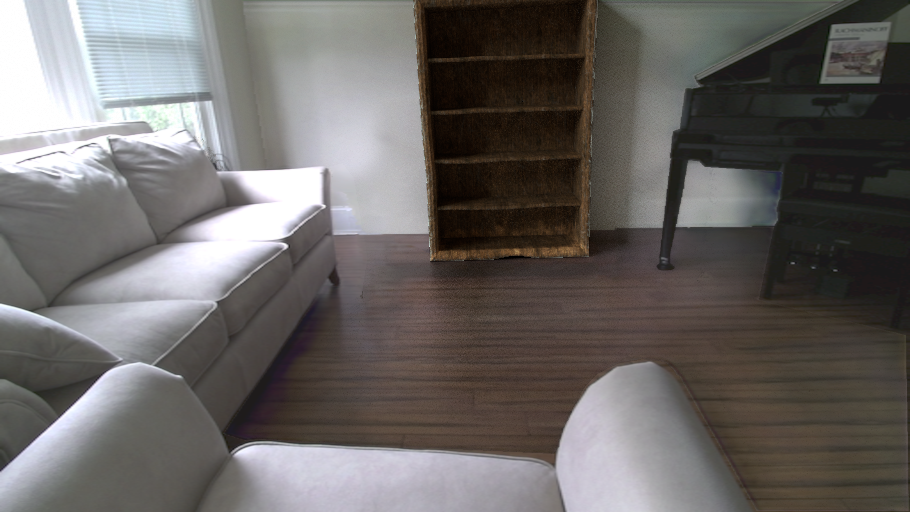}
    \caption{Virtual refurnishing applications, with input images on the left and refurnished results on the right.}
\label{fig:furniture}
\end{figure}

\subsection{Discussion and Future Work}
Our shadow removal enables much more realistic mixed reality
applications, ranging from consumer applications in real estate and furniture
retail, to socially beneficial uses for understanding physical resource
allocation in environments such as hospitals and schools. As an example,
Figures~\ref{fig:teaser}~and~\ref{fig:furniture} show results for a refurnishing
scenario. To generate these results, we run our pipeline twice -- once for the
wall, and once for the floor. We then composite the two results together, and
then insert a virtual object using standard differential rendering.
In the second row of Figure~\ref{fig:furniture}, the glossy hardwood floor shows a specularity of
the couch we wish to remove; by adding a specular component to the proxy model's floor plane, our
pipeline is able to remove the specularity as well as the shadow.

Of course, this specularity is fairly simple, where the couch ``occludes''
the reflection of the much brighter white wall.
Since glossy surfaces are
present in many indoor scenes, handling more complex specularities is an important area for further investigation.
In this vein, handling higher order light transport effects, such as color bleeding,
is also important for realistic results. Finally, in this work, we focused on single planar receivers, and did not
consider the shadows cast on other the objects in the scene. While our setup does
not depend on the planarity of the receiver,
the shadow receiver needed to have a consistent appearance in order to obtain good results with
complex textures. Extending the pipeline to handle multiple receivers, or
using different representations of textures, would be important additions to our
work.

As a final note, any digital image manipulation method carries the risk of misuse. This is especially true with the current prevalence of social media,
where false images may be used to spread misinformation and disinformation widely and rapidly. We
strongly believe in the importance of research on watermarking and other methods
to verify the authenticity of images and track image manipulations.

{\small
\bibliographystyle{ieee_fullname}
\bibliography{egbib}
}

\end{document}